\documentclass{ieeeaccess}
\usepackage{cite}
\usepackage{amsmath,amssymb,amsfonts}
\usepackage{multirow}

\usepackage{algorithm}
\usepackage{algpseudocode}
\usepackage{tablefootnote}
\usepackage{footnote}
\usepackage{caption}
\usepackage{graphicx}
\usepackage{textcomp}
\usepackage{hyperref}
\usepackage{color,soul}
\usepackage{tabularx}
\usepackage{ragged2e}
\def\BibTeX{{\rm B\kern-.05em{\sc i\kern-.025em b}\kern-.08em
    T\kern-.1667em\lower.7ex\hbox{E}\kern-.125emX}}
\begin{document}
\history{Date of publication xxxx 00, 0000, date of current version xxxx 00, 0000.}
\doi{10.1109/ACCESS}

\title{The Impact of LoRA Adapters on LLMs for Clinical Text Classification Under Computational and Data Constraints}

\author{\uppercase{Thanh-Dung Le}\authorrefmark{1,2},  \IEEEmembership{Senior Member, IEEE}, \uppercase{Ti Ti Nguyen}\authorrefmark{2}, \IEEEmembership{Member, IEEE}, \uppercase{Vu Nguyen Ha}\authorrefmark{2},  \IEEEmembership{Senior Member, IEEE}, \uppercase{Symeon Chatzinotas}\authorrefmark{2},  \IEEEmembership{Fellow, IEEE}, \uppercase{Philippe Jouvet}\authorrefmark{3}, \uppercase{and Rita Noumeir}\authorrefmark{1},
\IEEEmembership{Member, IEEE}}

\address[1]{Biomedical Information Processing Lab, \'{E}cole de Technologie Sup\'{e}rieure, Montr\'{e}al, Qu\'{e}bec, Canada}

\address[2]{The Interdisciplinary Centre for Security, Reliability, and Trust (SnT), University of Luxembourg, Luxembourg}
\address[3]{CHU Sainte-Justine Research Center, CHU Sainte-Justine Hospital, University of Montreal, Montr\'{e}al, Qu\'{e}bec, Canada}

\tfootnote{This work was supported in part by the Natural Sciences and Engineering Research Council (NSERC), in part by the Institut de Valorisation des données de l’Université de Montréal (IVADO), in part by the Fonds de la recherche en sante du Quebec (FRQS).}

\markboth
{Thanh-Dung Le \headeretal: The Impact of LoRA Adapters on LLMs for Clinical Text Classification Under Computational and Data Constraints}
{Thanh-Dung Le \headeretal: The Impact of LoRA Adapters on LLMs for Clinical Text Classification Under Computational and Data Constraints}

\corresp{Corresponding author: Thanh-Dung Le (e-mail: thanh-dung.le@uni.lu).}

\begin{abstract} 
Fine-tuning Large Language Models (LLMs) for clinical Natural Language Processing (NLP) poses significant challenges due to domain gap, limited data, and stringent hardware constraints. In this study, we evaluate four adapter techniques—Adapter, Lightweight, TinyAttention, and Gated Residual Network (GRN) - equivalent to Low-Rank Adaptation (LoRA), for clinical note classification under real-world, resource-constrained conditions. All experiments were conducted on a single NVIDIA Quadro P620 GPU (2 GB VRAM, 512 CUDA cores, 1.386 TFLOPS FP32), limiting batch sizes to $\leq$ 8 sequences and maximum sequence length to 256 tokens. Our clinical corpus comprises only 580 000 tokens, several orders of magnitude smaller than standard LLM pre-training datasets. We fine-tuned three biomedical pre-trained LLMs (CamemBERT-bio, AliBERT, DrBERT) and two lightweight Transformer models trained from scratch. Results show that (i) adapter structures provide no consistent gains when fine-tuning biomedical LLMs under these constraints, and (ii) simpler Transformers, with minimal parameter counts and training times under 6 hours, outperform adapter-augmented LLMs, which required over 1000 GPU-hours. Among adapters, GRN achieved the best metrics (accuracy, precision, recall, F1 = 0.88). These findings demonstrate that, in low-resource clinical settings with limited data and compute, lightweight Transformers trained from scratch offer a more practical and efficient solution than large LLMs, while GRN remains a viable adapter choice when minimal adaptation is needed.
\end{abstract}


\titlepgskip=-21pt

\maketitle

\section{Introduction}
\label{sec:introduction}
\PARstart{C}{urrently}, LLMs in natural language processing (NLP) have achieved remarkable advancements, evolving significantly over recent years. Before 2017, Long Short-Term Memory Networks (LSTMs) were the state-of-the-art in language modeling, reaching impressive scales of up to a billion parameters \cite{jozefowicz2016exploring}. The introduction of the Transformer model in 2017 marked a paradigm shift, leveraging the attention mechanism to set new benchmarks in NLP \cite{vaswani2017attention}. This innovation laid the groundwork for models such as GPT-2 \cite{radford2019language} and GPT-3 \cite{brown2020language}, and further studies into the scaling laws for neural language models \cite{kaplan2020scaling}. Today, Transformer-based architectures with self-attention mechanisms, exemplified by models like GPT-4, Claude 3, and Gemini, have become the standard for LLMs \cite{achiam2023gpt}.

\begin{figure}[!htp]
	\centering
	\includegraphics[scale=0.55]{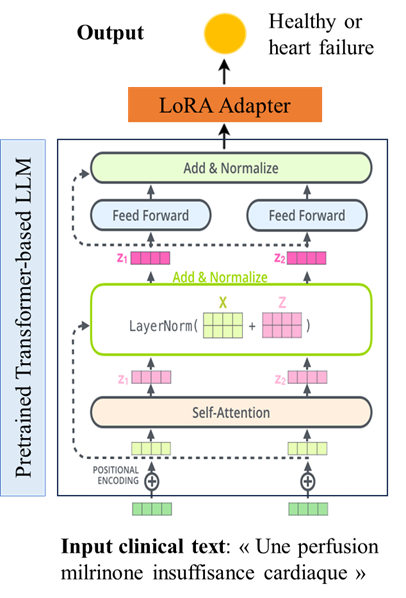}
	\caption{The visualized workflow for the experiment set-up with Transformer-based LLM \cite{alammar2018illustrated} structure, and learnable layers from a LoRA Adapter, which is a lightweight auxiliary network that runs alongside the transformer layer, transforming their activations into a structured, hierarchical feature representation.}
	\label{fig:llm_trend}
\end{figure}

In the clinical domain, the decision support system (CDSS) at CHU Sainte-Justine (CHUSJ) aims to enhance the diagnosis and management of acute respiratory distress syndrome (ARDS) in real-time by automatically analyzing data from electronic medical records, chest X-rays, and other sources. Previous research has highlighted that ARDS is often diagnosed late or missed in many patients, underscoring the need for more effective diagnostic tools \cite{bellani2016epidemiology}. Diagnosing ARDS requires identifying three main conditions: hypoxemia, chest X-ray infiltrates, and the absence of cardiac failure \cite{pediatric2015pediatric}. Furthermore, ARDS and cardiac failure frequently present with similar symptoms, making early and accurate diagnosis crucial for effective treatment strategies, particularly in critical care units like the Pediatric Intensive Care Unit (PICU). Accurately distinguishing between these conditions can significantly influence patient outcomes, potentially saving lives.

The research team at CHUSJ has developed advanced algorithms to detect hypoxemia \cite{sauthier2021estimated}, analyze chest X-rays \cite{zaglam2014computer, yahyatabar2020dense}, and identify the absence of cardiac failure. Our research group has also extensively analyzed machine learning algorithms for detecting cardiac failure from clinical narratives using NLP techniques \cite{le2021detecting, le2023adaptation}. Recent studies have demonstrated the superior performance of LLMs in handling complex tasks, such as understanding numerical attributes within clinical notes that contribute to cardiac failure, compared to traditional word embedding and deep learning methods \cite{lompo2024numerical, lompo2024multi}. Implementing these advanced algorithms has the potential to significantly increase ARDS diagnosis rates and improve patient outcomes at CHUSJ.

However, while efforts have been made to adapt LLMs in these studies, the results have been limited, indicating the need for further research and optimization to leverage LLM capabilities fully. Applying LLMs in clinical NLP remains challenging due to limited data availability and strict privacy regulations. Training must often be confined to protected environments within hospital servers, especially in CDSS environments that operate under constrained computational resources and inflexible data privacy policies. Despite promising results, these significant limitations persist. Consequently, this study empirically analyzes LLMs' adaptability within the CDSS framework at CHUSJ, aiming to enhance clinical decision-making and patient outcomes while navigating the challenges of data privacy and resource constraints.

In summary, as shown in Fig. \ref{fig:llm_trend}, this study addresses the challenge of adapting LLMs for clinical note classification within the strict data, privacy, and compute constraints of the CHUSJ CDSS. Our primary objectives are to (i) empirically evaluate lightweight adapter structures (Adapter, Lightweight, TinyAttention, GRN) for fine-tuning pre-trained biomedical LLMs under these constraints, (ii) benchmark their performance against Transformer models trained from scratch on a limited 580 000-token corpus, and (iii) derive practical recommendations for deploying NLP models in resource-limited clinical settings. 
The main contributions of this work are:
\begin{itemize}  
  \item We conduct the first head-to-head comparison of four LoRA-equivalent adapter techniques on three biomedical LLMs (CamemBERT-bio, AliBERT, DrBERT) versus lightweight Transformers trained from scratch.
 \item We identify GRN as the top-performing adapter (F1 = 0.88) and demonstrate that simpler Transformers reach superior accuracy in under 6 GPU-hours.
\end{itemize}  

The remainder of the paper is organized as follows. Section \ref{sec:related_works} reviews related work on adapter methods and clinical NLP. Section \ref{sec:dataset_method} describes our dataset, experimental setup, and adapter architectures. Section \ref{sec:result_discussions} presents quantitative results and analysis. Section \ref{sec:limitations} discusses limitations and implications for CDSS integration and deployment. Finally, Section \ref{sec:conclusion} concludes and outlines future directions.

\section{Related Works}
\label{sec:related_works}
One of the critical challenges with Transformer-based LLMs in clinical text classification is their difficulty in accurately interpreting short texts and their tendency to rely heavily on keywords \cite{le2021detecting}. In our recent research, we have explored various strategies to improve LLM performance in this domain. These strategies include utilizing Mixture of Experts (MoE) Transformers \cite{le2023small} and integrating adapters as intermediate layers to filter out irrelevant information \cite{le2023grn}. Despite these efforts, these approaches did not surpass the performance of a simple MLP combined with a dense feature representation from an autoencoder \cite{le2023adaptation}. This underperformance is attributed to a generalization gap between training and validation, especially with large models trained on small datasets. Additionally, other findings indicate that LLMs may not consistently deliver superior results, particularly when considering accuracy, cost, and safety factors. As models become more complex and expensive, issues related to cost and accessibility become more pronounced, which are critical factors in the CDSS environment \cite{fields2024survey, raiaan2024review}.

Several potential approaches can be employed to address the challenges of using LLMs in clinical NLP with small, limited datasets. One effective strategy is instruction tuning, a parameter-efficient method that optimizes LLMs to follow specific instructions better, thereby aligning them to new domains \cite{singhal2023large}. Additionally, fine-tuning techniques can help unlock the capabilities of LLMs for various downstream applications, ensuring robust performance even with constrained data \cite{zhangscaling}. These strategies can significantly enhance the adaptability and effectiveness of LLMs in clinical settings where data availability is limited, ultimately improving their utility.

Two primary approaches are commonly employed in fine-tuning LLMs: full-model tuning (FMT) and parameter-efficient tuning (PET). PET includes methods such as prompt tuning and LoRA, which are especially relevant when the size of the LLM far exceeds the available fine-tuning data, a common scenario in data-limited environments \cite{chen2022revisiting}. Among these methods, LoRA is particularly notable due to its adaptability and ability to facilitate end-to-end customization during fine-tuning. LoRA freezes the pre-trained model weights and introduces trainable rank decomposition matrices into each layer of the Transformer architecture. This significantly reduces the number of trainable parameters required for downstream tasks. For example, compared to GPT-3 175B fine-tuned with Adam, LoRA can reduce the number of trainable parameters by 10,000 times and the GPU memory requirement by three times. LoRA performs as well as or better than traditional fine-tuning in terms of model quality on models such as RoBERTa, DeBERTa, GPT-2, and GPT-3, despite having fewer trainable parameters, higher training throughput, and no additional inference latency \cite{hulora}.

Adapter modules \cite{houlsby2019parameter, pfeiffer2020adapterhub} represent a form of LoRA efficient tuning, integrating small, newly initialized parameter modules at each transformer layer of pre-trained LLMs. These modules typically comprise a two-layer feed-forward neural network with a bottleneck structure. Specifically, the adapter structure includes (1) a down-projection layer with weights $W_{down} \in \mathbb{R}^{d \times r}$ that reduces the input $h_i$ to a lower-dimensional space defined by the bottleneck dimension $r$; and (2) an up-projection layer with weights $W_{up} \in \mathbb{R}^{r \times d}$ that projects the reduced input back to its original size. Mathematically, the adapter operation can be expressed as:

\begin{equation}
h_a = W_{up}^T f \left( W_{down}^T h_i \right)
\end{equation}

where $h_a$ is the output and $f(\cdot)$ represents the activation function. This configuration allows for efficient parameter updates during fine-tuning while maintaining the overall structure and performance of the pre-trained LLMs.
Therefore, this study aims to analyze the impact of different adapter structures, which offer minimal complexity and rapid adaptation to LLMs, for clinical NLP narrative classification. This implementation is designed to operate within constrained computational capacities, making it suitable for environments with limited computational resources. The choice of NVIDIA Quadro P620, with its significantly limited computational capabilities (512 CUDA cores and only 1.386 TFLOPS FP32 performance), imposes substantial computational constraints compared to high-performance GPUs such as the NVIDIA A100, as shown in Table \ref{tab:nvidia_gpu}. This selection reflects deliberate experimental conditions intended to replicate scenarios typical of resource-constrained environments, ensuring that developed models are robust and efficient under strict hardware limitations for fine-tuning LLMs directly on clinical texts.

\begin{figure*}[!htp]
	\centering
	\includegraphics[scale=0.5]{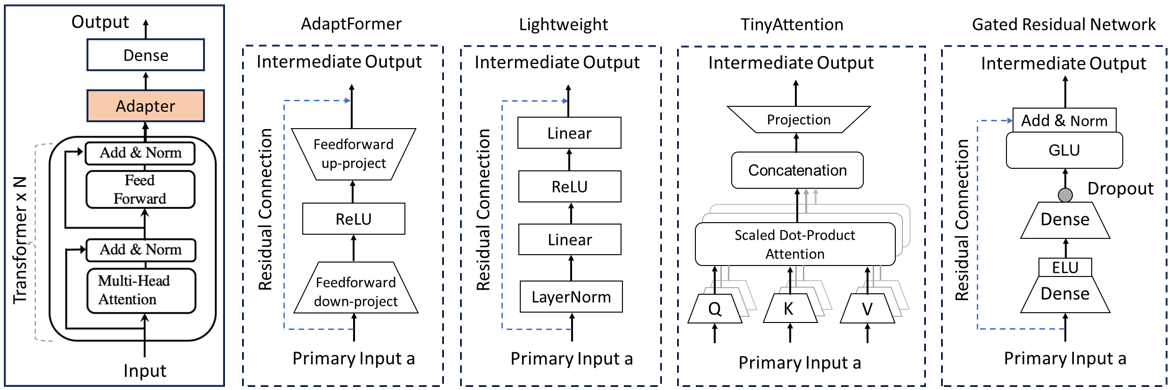}
	\caption{Different adapter structure, including AdaptFormer \cite{chen2022adaptformer}, Lightweight \cite{le2021lightweight}, TinyAttention \cite{zhao2022tiny}, GRN \cite{le2023grn, le2024transformer}}
	\label{fig:adapter_structure}
\end{figure*}

\begin{table*}[ht]
\centering
\caption{Comparison between NVIDIA Quadro P620 and NVIDIA A100 80GB PCIe}
\begin{tabular}{|l|l|l|}
\hline
\textbf{Specifications} & \textbf{NVIDIA Quadro P620} & \textbf{NVIDIA A100 80GB PCIe} \\ \hline
Architecture & Pascal (GP108) & Ampere (GA100) \\ \hline
CUDA Cores & 512 & 6,912 \\ \hline
Tensor Cores & n/a & 432 (3rd-gen, TF32 \& FP64 support) \\ \hline
Memory & 2\,GB GDDR5 & 80\,GB HBM2e \\ \hline
Computing Performance & \begin{tabular}[t]{@{}l@{}}
• FP32: 1.386 TFLOPS\\

\end{tabular} & 
\begin{tabular}[t]{@{}l@{}}
• FP32: 19.5 TFLOPS\\
• FP16 (Tensor): 312 TFLOPS (624 w/sparsity)\\
• INT8 (Tensor): 624 TOPS (1,248 w/sparsity)
\end{tabular} \\ \hline 
Memory Bus / Bandwidth & 128-bit / 48\,GB/s & – / 2,039\,GB/s \\ \hline
Thermal Design Power (TDP) & 40\,W & 250\,W \\ \hline
Key Strengths & 
\begin{tabular}[t]{@{}l@{}}
• Small, power-efficient (40\,W TDP)\\
\end{tabular} & 
\begin{tabular}[t]{@{}l@{}}
• Massive Tensor performance (~312 TFLOPS)\\
• Multi-Instance GPU (up to 7 instances)\\
• High PCIe 4.0 bandwidth
\end{tabular} \\ \hline
\end{tabular}
\label{tab:nvidia_gpu}
\end{table*}

\section{Materials and Methods}
\label{sec:dataset_method}

\subsection{Clinical Notes Data at CHUSJ}
This study was conducted following ethical approval from the research ethics board at CHUSJ (protocol number: 2020-2253), and the study's design focused on identifying cardiac failure in patients within the first 24 hours of admission by analyzing admission and evolution notes during this initial period. The dataset consisted of 580,000 unigrams extracted from 5,444 single lines of short clinical narratives. Of these, 1,941 cases were positive (36\% of the total), and 3,503 cases were negative. While the longest n-gram was over 400 words, most n-grams had a length distribution between 50 and 125 words. The average length of the number of characters was 601 and 704, and the average size of the number of digits was 25 and 26 for the positive and negative cases, respectively. We pre-processed the data by removing stop-words and accounting for negation in medical expressions. Numeric values for vital signs (heart rate, blood pressure, etc.) were also included and decoded to account for nearly 4\% of the notes containing these values. All notes are short narratives; detailed characteristics for the notes at CHUSJ can be found in the Supplementary Materials from the study \cite{le2021detecting, le2023adaptation}.

\begin{figure*}[!pt]
	\centering
	\includegraphics[angle=90, width=\linewidth,height=\textheight,keepaspectratio]{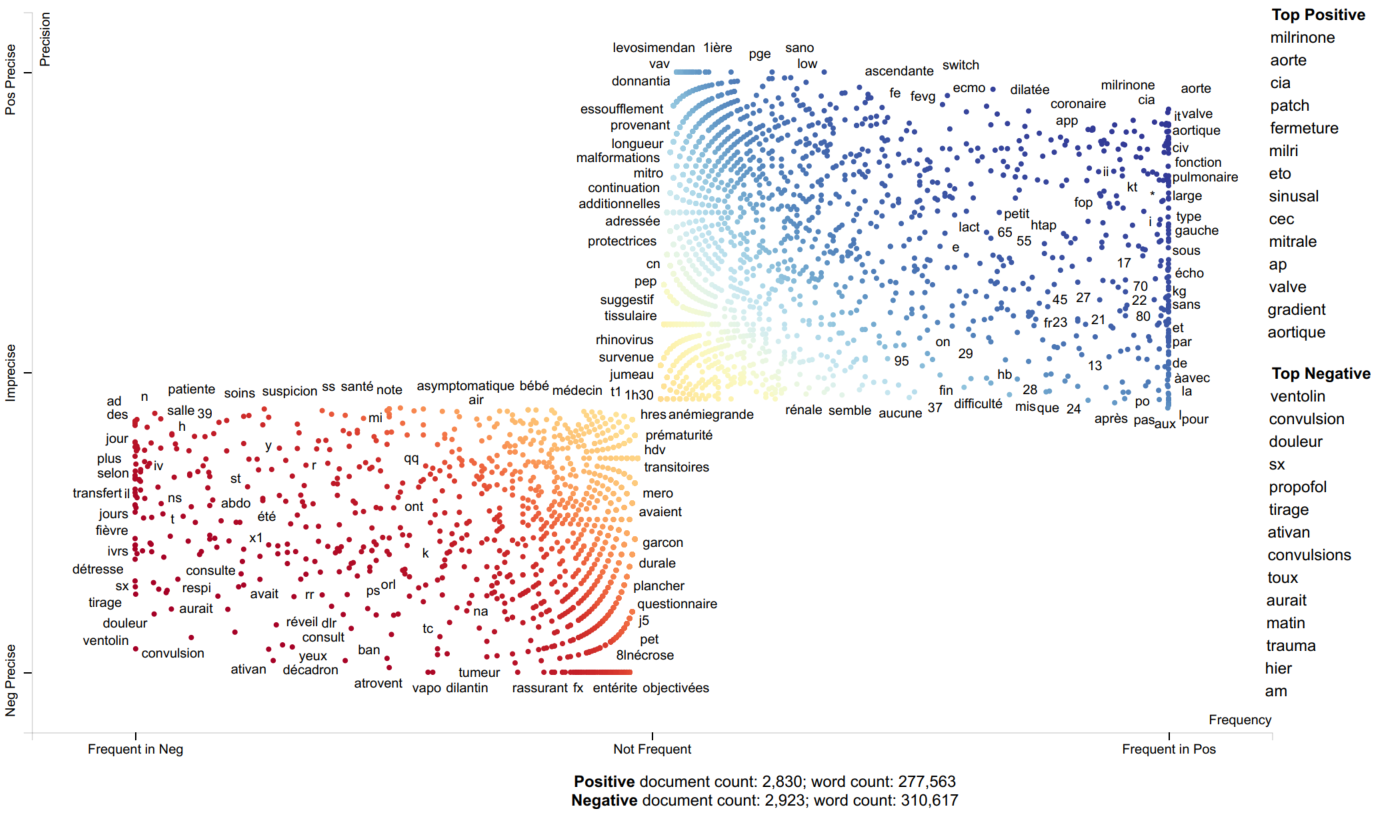}
	\caption{Overview of word distribution in clinical notes at CHUSJ \cite{le2021detecting}.}
	\label{fig:clinical_nlp_illustration}
\end{figure*}

In summarization, we apply the ScatterText \cite{kessler2017scattertext} for the note visualization. In total, we have over 580000 unigrams (n-gram) shown in Fig. \ref{fig:clinical_nlp_illustration}. The figure shows the most frequent words for the positive case in the upper right corner; the most frequent words for the negative cases in the lower-left corner; and, all less frequent words for both cases are in the center. Besides, the top terms from the positive and negative cases are presented on the right-hand side. In positive cases, we quickly see that most of these terms are positively related to cardiac malfunction: milrinone or milri (milrinone), aorte or aortique valve (aortic valve). In contrast, terms such as respiratoire (respiratory), and IVRS (Infection des voies respiratoires supérieures - Virus responsible for respiratory distress) indicate respiratory syndromes. 

\subsection{Biomedical Pretrained LLM}
In recent years, the development of biomedical pre-trained BERT-based models has significantly advanced the processing and understanding of biomedical text, particularly within the French language domain. As confirmed in \cite{le2023small}, three notable among these models are CamemBERT-bio \cite{corr2306}, DrBERT \cite{LabrakBDRMDG23}, and AliBERT \cite{BerheDMMDZ23}, each tailored to address the unique challenges of biomedical text analysis. CamemBERT-bio, for instance, is designed explicitly for French biomedical data, leveraging the robust architecture of CamemBERT to deliver superior performance in this field. Similarly, DrBERT and AliBERT enhance the landscape of specialized models by offering high accuracy and efficiency in various biomedical NLP tasks. These models are exceptionally well-suited for classifying French clinical notes, having been trained on extensive French biomedical corpora. These models are particularly adept at classifying French clinical notes due to their training in extensive French biomedical corpora, which enables them to accurately capture the nuances and specialized terminology unique to French medical practice.

\subsection{Transformer-based Models}
Training Transformer models effectively with small datasets presents a significant challenge. Transformers often exhibit limitations such as a generalization gap and sharp minima when applied to small datasets \cite{le2021detecting}. Furthermore, their performance degrades on imbalanced and small clinical datasets \cite{macabiau2024label}. Our recent study indicates that the Mixture-of-Experts (MoE) Transformer \cite{le2023small} can mitigate some of these limitations by enhancing model performance with limited data. In this study, we will experiment with the standard Transformer and the MoE-Transformer for clinical text classification tasks to evaluate their effectiveness in handling small and imbalanced datasets.

\begin{table*}[ht]
\centering
\footnotesize
\caption{Comparison of adapters for Transformer with Computational Complexity}
\begin{tabular}{|m{2.5cm}|m{4cm}|m{5cm}|m{4cm}|}
\hline
\textbf{Models} & \textbf{Structure Summarization} & \textbf{Complexity (FLOPs)} & \textbf{Highlights} \\
\hline
\textbf{AdaptFormer} & 
\begin{itemize}
    \item Feedforward down-project: $M \rightarrow D$
    \item ReLU: $D$
    \item Feedforward up-project: $D \rightarrow M$
\end{itemize} & $4 \cdot M \cdot D + D$ & Least complex if $D$ is much smaller than $M$ \\
\hline
\textbf{Lightweight} & 
\begin{itemize}
    \item LayerNorm: $M$
    \item Linear: $M \rightarrow D$
    \item ReLU: $D$
    \item Linear: $D \rightarrow M$
\end{itemize} & $4 \cdot M \cdot D + D + 2 \cdot M$ & Dominated by $4 \cdot M \cdot D$; additional small terms \\
\hline
\textbf{TinyAttention} & 
\begin{itemize}
    \item Q, K, V projections: $M \rightarrow \frac{D}{h}$
    \item Scaled dot-product attention
    \item Concatenation
    \item Projection: $D$
\end{itemize} & $6 \cdot \frac{M \cdot D}{h} + 4 \cdot N \cdot D + \frac{2 \cdot N^2 \cdot D}{h} + 2 \cdot M \cdot D$ & Complexity highly dependent on $h$ and $N$ \\
\hline
\textbf{GRN} & 
\begin{itemize}
    \item Dense: $M \rightarrow D$
    \item ELU: $D$
    \item Dense: $D \rightarrow M$
    \item Dropout: $D$
    \item GLU: Element-wise multiplication (gate)
    \item Add \& Norm: $M$
\end{itemize} & $4 \cdot M \cdot D + 4 \cdot D + 2 \cdot M$ & Similar leading term as Lightweight; slightly higher additional terms \\
\hline
\end{tabular}
\label{tab:adapter-comparison}
\end{table*}

\subsection{Adapters Structures}
Employing limited computational constraints and limited data, this study limited the experiment to the following adapter structure, which is simple and scalable for effectively fine-tuning the pre-trained model as the LoRA technique \cite{ding2023parameter}, as shown in Fig. \ref{fig:adapter_structure}. Below we present the formal derivations for each adapter, specifying how an input token embedding \(\mathbf{x}\in\mathbb{R}^d\) is transformed into an output \(\mathbf{y}\in\mathbb{R}^d\).

\subsubsection{AdaptFormer \cite{chen2022adaptformer}}
AdaptFormer is a parameter-efficient tuning module for Transformer architectures that enhances adaptability by incorporating a feedforward down-projection layer, a ReLU activation, and an up-projection layer to restore input size. It includes a residual connection to preserve the original input, improving learning without significantly increasing model complexity. By updating only the adapter modules' parameters, AdaptFormer enables effective fine-tuning while keeping the pre-trained model fixed.

\begin{align}
  \mathbf{h}_{\downarrow} &= W_{\downarrow}\,\mathbf{x},\quad W_{\downarrow}\in\mathbb{R}^{r\times d},\\
  \mathbf{h}_{\text{act}}   &= \phi(\mathbf{h}_{\downarrow}),\quad \phi\in\{\mathrm{ReLU},\mathrm{GELU}\},\\
  \mathbf{h}_{\uparrow}   &= W_{\uparrow}\,\mathbf{h}_{\text{act}},\quad W_{\uparrow}\in\mathbb{R}^{d\times r},\\
  \mathbf{y}               &= \mathbf{x} + \mathbf{h}_{\uparrow}.
\end{align}

\subsubsection{Lightweight \cite{le2021lightweight}}
The Lightweight adapter structure integrates a linear down-projection layer followed by a ReLU activation, a second linear layer, and a final LayerNorm for normalization. This configuration is enhanced with a residual connection to maintain the original input alongside the processed output. By focusing on linear transformations and normalization, this adapter efficiently fine-tunes the model with minimal additional parameters, ensuring lightweight adaptability.

\begin{align}
    \mathbf{y} = \mathbf{x} + W_{\ell}\,\mathbf{x},\quad W_{\ell}\in\mathbb{R}^{d\times r}
\end{align}

\subsubsection{TinyAttention \cite{zhao2022tiny}}
The TinyAttention adapter structure incorporates scaled dot-product attention, where the primary input is split into query (Q), key (K), and value (V) components. The attention mechanism calculates attention weights and produces a weighted sum of the values, which are then concatenated and passed through a projection layer. This structure allows the model to focus on relevant input parts efficiently, enhancing the representation with minimal additional parameters.

\begin{align}
  Q' &= W_Q\,\mathbf{x},\;W_Q\in\mathbb{R}^{r\times d},\\
  K' &= W_K\,\mathbf{x},\;W_K\in\mathbb{R}^{r\times d},\\
  V' &= W_V\,\mathbf{x},\;W_V\in\mathbb{R}^{r\times d},\\
  A  &= \text{softmax}\bigl(Q'K'^\top/\sqrt{r}\bigr)\,V',\\
  \mathbf{y} &= \mathbf{x} + W_O\,A, \quad W_O\in\mathbb{R}^{d\times r}
\end{align}

\subsubsection{Gated Residual Networks (GRN) \cite{le2023grn, le2024transformer}}

The GRN adapter structure includes a series of dense layers. The primary input is first processed through an ELU activation function and a dense layer. The output then passes through a dropout layer and another dense layer before being gated by a gated linear unit (GLU), $\odot$ is the element-wise Hadamard product. Finally, the gated output is added to the original input via a residual connection, followed by normalization (Add \& Norm), enhancing the model's ability to learn complex representations efficiently while maintaining stability.

\begin{align}
  \mathbf{u} &= \mathrm{ELU}\bigl(W_1\,\mathbf{x}\bigr),\quad W_1\in\mathbb{R}^{r\times d},\\
  \mathbf{v} &= W_2\,\mathbf{u},\quad W_2\in\mathbb{R}^{d\times r},\\
  g          &= \sigma\bigl(W_g\,\mathbf{v}\bigr),\quad W_g\in\mathbb{R}^{d\times d},\\
  \mathbf{h} &= g\odot \mathbf{v},\\
  \mathbf{y} &= \mathrm{LayerNorm}( \mathbf{x}+ \mathbf{h})
\end{align}

For the complexity of each adapter structure, Table \ref{tab:adapter-comparison} compares the computational complexity of different adapter structures for Transformers. AdaptFormer, with its simple feedforward layers and ReLU activation, has the least complexity when the bottleneck dimension \(D\) is significantly smaller than the model dimension \(M\). The Lightweight adapter adds LayerNorm and utilizes linear transformations, resulting in a complexity dominated by \(4 \cdot M \cdot D\). TinyAttention introduces attention mechanisms, making its complexity highly dependent on the number of heads \(h\) and sequence length \(N\). The GRN includes dense layers, ELU activation, dropout, and a gated linear unit, leading to a complexity similar to the Lightweight adapter but with slightly higher additional terms. These approaches are particularly suitable for fine-tuning a pre-trained LLM on a limited dataset and under constrained computational capacity.

\begin{figure}[!htp]
	\centering
	\includegraphics[scale=0.35]{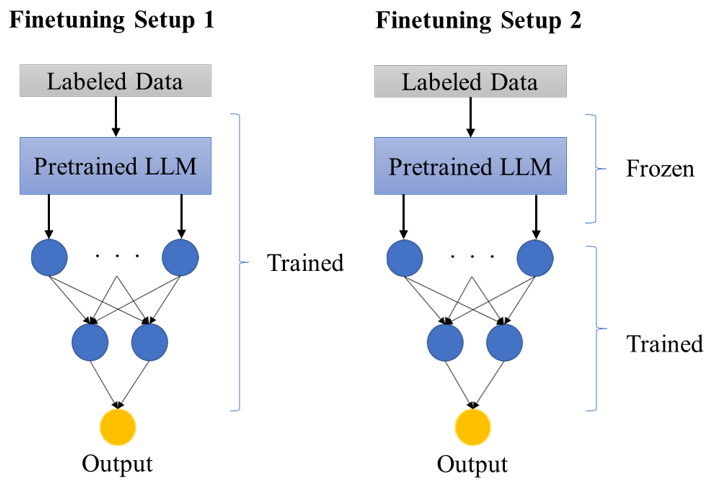}
	\caption{Experiment setup}
	\label{fig:setup}
\end{figure}

\section{Experimental Results}
\label{sec:result_discussions}

We employed two fine-tuning approaches for the experimental setup, as illustrated in Figure \ref{fig:setup}. In \textbf{Fine-tuning Setup 1}, the entire pre-trained language model (LLM) is fine-tuned using labeled data, where both the pre-trained LLM and the additional layers are trained simultaneously. This method allows the model to fully adapt to the specific task by updating all parameters, potentially leading to better performance, but it is computationally intensive. In \textbf{Fine-tuning Setup 2}, the pre-trained LLM is frozen, meaning its parameters are not updated during fine-tuning, and only the additional layers appended to the LLM are trained using labeled data. This approach reduces computational requirements and mitigates the risk of overfitting, making it more suitable for scenarios with limited data and computational resources. By comparing these setups, we aim to evaluate the effectiveness and efficiency of fine-tuning strategies for clinical text classification tasks.

Table \ref{tab:specification_adapter} compares four models: AdaptFormer, Lightweight, GRN, and TinyAttention. For each model, it outlines its specifications and the total number of parameters, including the memory footprint in megabytes (MB). AdaptFormer uses a down projection dimension of 512 and an up projection dimension of 1024, totaling 1,839,618 parameters (7.02 MB). The Lightweight model has an input dimension of 1024 with 2,890,754 parameters (11.03 MB). GRN features an input dimension 1024 with a drop-out rate of 0.5, amounting to 3,940,354 parameters (15.03 MB). Finally, TinyAttention, which includes an input dimension of 1024, four heads, and a drop-out rate of 0.25, has the highest number of parameters at 9,188,354 (35.05 MB).

\begin{table*}[h!]
\small
\centering
\caption{Model Specifications and Total Parameters for Adapter Structures}
\begin{tabular}{|l|l|l|}
\hline
Model & Specifications & Total parameters \\ \hline
AdaptFormer & Down projection dimension 512; Up projection dimension 1024 & 1,839,618 (7.02 MB) \\ \hline
Lightweight & Input dimension 1024 & 2,890,754 (11.03 MB) \\ \hline
GRN & Input dimension 1024, drop-out at 0.5 & 3,940,354 (15.03 MB) \\ \hline
TinyAttention & Input dimension 1024, 4 heads, and drop-out at 0.25 & 9,188,354 (35.05 MB) \\ \hline
\end{tabular}
\label{tab:specification_adapter}
\vspace{3mm}
\end{table*}

\begin{table*}[h!]
\centering
\footnotesize
\caption{Performance Comparison of CamemBERT-bio with Different Adapters.} 
\begin{tabular}{|c|c|c|c|c|c|c|}
\hline
CamemBERT-bio & Acc $(\uparrow)$ & Pre $(\uparrow)$ & Rec $(\uparrow)$ & F1 $(\uparrow)$ & Training Time (hours) ($\downarrow$) & Inference Time (s) ($\downarrow$) \\
\hline
\multicolumn{7}{|c|}{Setup 1} \\
\hline
Baseline & \textbf{0.87} & 0.86 & 0.88 & \textbf{0.87} & 31.7 & 142 \\
Adapter & 0.84 & 0.85 & 0.83 & 0.84 & 22.5 & 130 \\
Lightweight & 0.84 & 0.83 & 0.82 & 0.82 & 30.8 & 121 \\
GRN & 0.86 & 0.83 & \textbf{0.91} & 0.87 & 27 & 131 \\
TinyAttention & 0.83 & 0.79 & 0.88 & 0.83 & 23 & 122 \\
\hline
\multicolumn{7}{|c|}{Setup 2} \\
\hline
Adapter & 0.74 & 0.78 & 0.72 & 0.75 & 22.5 & 130 \\
Lightweight & 0.72 & 0.71 & 0.72 & 0.71 & 35.9 & 126 \\
GRN & 0.71 & 0.7 & 0.76 & 0.73 & 54.1 & 123 \\
TinyAttention & 0.71 & 0.7 & 0.72 & 0.71 & 41.6 & 127 \\
\hline
\end{tabular}
\label{tab:camemBERT}
\vspace{1mm} 
\captionsetup{font=footnotesize, justification=raggedright, singlelinecheck=false} 
\caption*{\hspace{20mm}\textbf{Bold} denotes the best values.}
\end{table*}

\begin{table*}[h!]
\centering
\footnotesize
\caption{Performance Comparison of AliBERT with Different Adapters.}
\begin{tabular}{|c|c|c|c|c|c|c|}
\hline
AliBERT & Acc $(\uparrow)$ & Pre $(\uparrow)$ & Rec $(\uparrow)$ & F1 $(\uparrow)$ & Training Time (hours) ($\downarrow$) & Inference Time (s) ($\downarrow$) \\
\hline
\multicolumn{7}{|c|}{Setup 1} \\
\hline
Baseline & 0.86 & \textbf{0.87} & 0.84 & 0.86 & 39.3 & 128 \\
Adapter & 0.78 & 0.72 & \textbf{0.88} & 0.79 & 42.8 & 128 \\
Lightweight & 0.84 & 0.82 & 0.85 & 0.84 & 34.8 & 131 \\
GRN & \textbf{0.87} & 0.84 & 0.84 & \textbf{0.85} & 30.8 & 127 \\
TinyAttention & 0.84 & 0.81 & 0.83 & 0.82 & 42.2 & 128 \\
\hline
\multicolumn{7}{|c|}{Setup 2} \\
\hline
Adapter & 0.68 & 0.67 & 0.67 & 0.67 & 46.6 & 126 \\
Lightweight & 0.66 & 0.72 & 0.49 & 0.58 & 46.7 & 126 \\
GRN & 0.67 & 0.7 & 0.59 & 0.64 & 37.5 & 126 \\
TinyAttention & 0.67 & 0.67 & \textbf{0.7} & 0.68 & 46.2 & 130 \\
\hline
\end{tabular}
\label{tab:alibert}
\vspace{1mm} 
\captionsetup{font=footnotesize, justification=raggedright, singlelinecheck=false} 
\caption*{\hspace{22mm}\textbf{Bold} denotes the best values.}
\end{table*}

\begin{table*}[h!]
\centering
\footnotesize
\caption{Performance Comparison of DrBERT with Different Adapters.}
\begin{tabular}{|c|c|c|c|c|c|c|}
\hline
DrBERT & Acc $(\uparrow)$ & Pre $(\uparrow)$ & Rec $(\uparrow)$ & F1 $(\uparrow)$ & Training Time (hours) ($\downarrow$) & Inference Time (s) ($\downarrow$) \\
\hline
\multicolumn{7}{|c|}{Setup 1} \\
\hline
Baseline & \textbf{0.87} & 0.84 & \textbf{0.9} & \textbf{0.87} & 45.2 & 133 \\
Adapter & 0.86 & \textbf{0.87} & 0.87 & 0.87 & 38.5 & 126 \\
Lightweight & 0.71 & 0.78 & 0.56 & 0.65 & 41.6 & 132 \\
GRN & 0.86 & 0.84 & 0.88 & 0.86 & 41.3 & 130 \\
TinyAttention & 0.85 & 0.81 & 0.9 & 0.85 & 28.9 & 123 \\
\hline
\multicolumn{7}{|c|}{Setup 2} \\
\hline
Adapter & 0.69 & 0.71 & 0.63 & 0.67 & 46.1 & 122 \\
Lightweight & 0.73 & 0.72 & 0.73 & 0.72 & 45 & 125 \\
GRN & 0.73 & 0.7 & 0.76 & 0.73 & 47 & 126 \\
TinyAttention & 0.75 & 0.76 & 0.69 & 0.72 & 47.5 & 123 \\
\hline
\end{tabular}
\label{tab:drbert}
\vspace{1mm} 
\captionsetup{font=footnotesize, justification=raggedright, singlelinecheck=false} 
\caption*{\hspace{22mm}\textbf{Bold} denotes the best values.}
\end{table*}

\begin{table*}[h!]
\centering
\footnotesize
\caption{Performance Comparison of Transformer with Different Adapters.}
\begin{tabular}{|c|c|c|c|c|c|c|}
\hline
Transformer & Acc $(\uparrow)$ & Pre $(\uparrow)$ & Rec $(\uparrow)$ & F1 $(\uparrow)$ & Training Time (hours) ($\downarrow$) & Inference Time (s) ($\downarrow$)  \\
\hline
Baseline & 0.85 & 0.85 & 0.83 & 0.84 & 0.11 & 3 \\
Adapter & 0.85 & 0.83 & 0.85 & 0.84 & 0.4 & 2 \\
Lightweight & 0.85 & 0.82 & 0.88 & 0.85 & 1 & 3 \\
GRN & \textbf{0.87} & 0.85 & \textbf{0.89} & \textbf{0.87} & 0.7 & 3 \\
TinyAttention & 0.85 & 0.81 & 0.88 & 0.84 & 0.7 & 3 \\
\hline
\end{tabular}
\label{tab:transformer}
\vspace{1mm} 
\captionsetup{font=footnotesize, justification=raggedright, singlelinecheck=false} 
\caption*{\hspace{22mm}\textbf{Bold} denotes the best values.}
\end{table*}

\begin{table*}[h!]
\centering
\footnotesize
\caption{Performance Comparison of MoE-Transformer with Different Adapters}
\begin{tabular}{|c|c|c|c|c|c|c|}
\hline
MoE-Transformer & Acc $(\uparrow)$ & Pre $(\uparrow)$ & Rec $(\uparrow)$ & F1 $(\uparrow)$ & Training Time (hours) ($\downarrow$) & Inference Time (s) ($\downarrow$)  \\
\hline
Baseline & 0.87 & 0.87 & 0.85 & 0.86 & 0.17 & 4 \\
Adapter & 0.84 & 0.78 & \textbf{0.92} & 0.84 & 0.4 & 2 \\
Lightweight & 0.84 & 0.76 & \textbf{0.95} & 0.84 & 1.2 & 4 \\
GRN & \textbf{0.88} & \textbf{0.88} & 0.88 & \textbf{0.88} & 0.8 & 3 \\
TinyAttention & 0.84 & 0.79 & 0.9 & 0.84 & 0.8 & 3 \\
\hline
\end{tabular}
\label{tab:moe_transformer}
\vspace{1mm} 
\captionsetup{font=footnotesize, justification=raggedright, singlelinecheck=false} 
\caption*{\hspace{20mm}\textbf{Bold} denotes the best values.}
\vspace{-5mm}
\end{table*}

All experiments were conducted on the Intel(R) Xeon(R) CPU E3-1225, 3.30GHz, 16GB RAM, and Nvidia Quadro P620 GPU, 2GB. For the implementation, experiments were implemented using the scikit-learn library \cite{scikit-learn}, and Keras \cite{chollet2015keras}. The data was divided into 70\% training and 30\% testing. Moreover, training and fine-tuning the Transformer-based model is complex. As reported by \cite{popel2018training}, model size, learning rate, batch size, and maximum sequence length are the four critical hyperparameters that significantly influence the training process of the Transformer model. In addition, we also applied dropout \cite{srivastava2014dropout} (p=0.25) and GlorotNormal kernel initializer \cite{glorot2010understanding}, batch normalization \cite{ioffe2015batch, bjorck2018understanding} are employed for models' stability. Additionally, we also apply early stopping based on the validation loss. Consequently, these hyperparameters were carefully chosen to achieve optimal performance and prevent overfitting.

To effectively assess the performance of our method, metrics including accuracy, precision, recall (or sensitivity), and F1 score were used \cite{goutte2005probabilistic}. These metrics are defined as follows: 
\begin{align}
&\text {Accuracy (acc) }=\frac{\mathrm{TP}+\mathrm{TN}}{\mathrm{TP}+\mathrm{TN}+\mathrm{FP}+\mathrm{FN}}  \\ 
&\text {Precision (pre) }=\frac{\mathrm{TP}}{\mathrm{TP}+\mathrm{FP}} \\
&\text {Recall/Sensitivity (rec)}=\frac{\mathrm{TP}}{\mathrm{TP}+\mathrm{FN}}  \\ 
&\text {F1-Score (f1)} =\frac{2^{\star} \text {Precision}^{\star} \text {Recall}}{\text {Precision }+\text {Recall}} 
\end{align}

\noindent where TN and TP stand for true negative and true positive, respectively, and they are the number of negative and positive patients that are classified correctly. Whereas FP and FN represent false positive and false negative, respectively, and they represent the number of positive and negative patients that were wrongly predicted. 

\begin{figure}[!htp]
	\centering
	\includegraphics[scale=0.5]{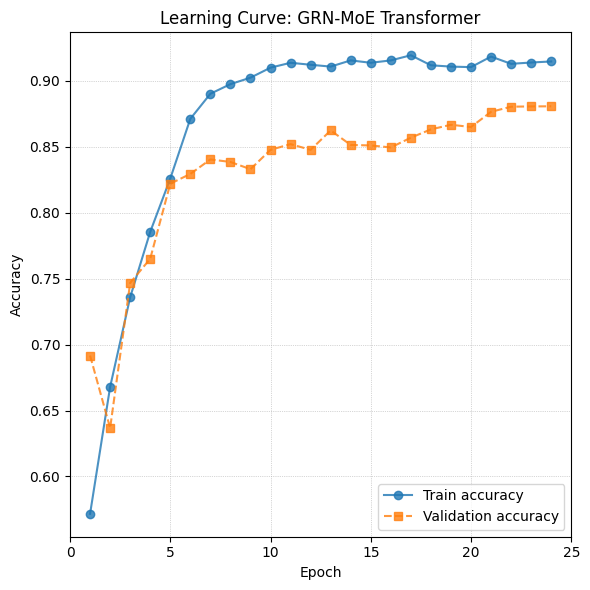}
    \includegraphics[scale=0.5]{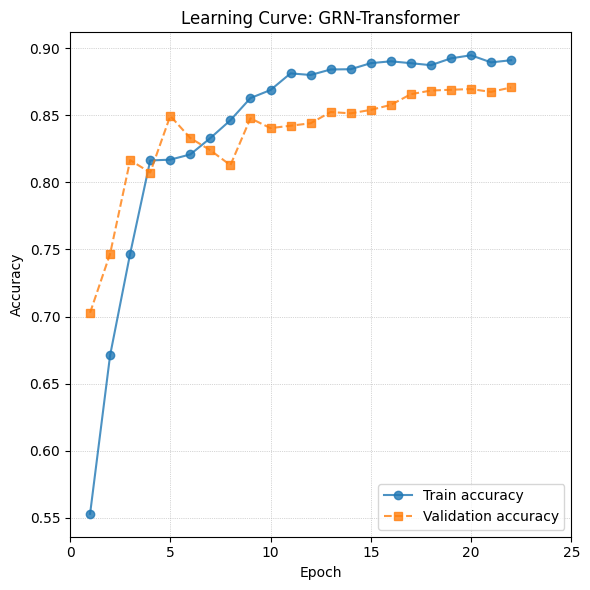}
	\caption{Learning curve performance of Transformer and MoE-Transformer with the GRN adapter.}
	\label{fig:learning_curve}
\end{figure}

As shown in Fig. \ref{fig:learning_curve}, both the GRN-Transformer and the GRN-MoE Transformer, trained from scratch, converge rapidly, exceeding 80\% accuracy by the 5th epoch, and maintain a narrow train–validation gap throughout. The MoE variant achieves a higher peak training accuracy ($\approx$ 91.5\% vs.\ 89\%) and reaches $\approx$ 88\% validation accuracy by epoch 24, indicating that expert routing effectively regularizes training. These results demonstrate that both architectures deliver strong predictive performance with minimal overfitting, making them well-suited for resource-constrained clinical tasks.

First of all, the experimental results compare the performance of various adapters applied to biomedical pre-trained LLMs (CamemBERT-bio, AliBERT, DrBERT) and Transformer-based models trained from scratch, evaluated on accuracy, precision, recall, F1 score, training time, and inference time as summarized in Table \ref{tab:camemBERT} to \ref{tab:moe_transformer}, respectively. In Setup 1, where full fine-tuning was applied, the baseline models achieved the highest performance across most metrics, with the GRN adapter showing competitive results. However, in Setup 2, where pre-trained weights were frozen, and only the adapters were fine-tuned, there was a significant performance decline across all adapters, demonstrating lower accuracy, precision, recall, and F1 scores. Notably, each experiment with biomedical pre-trained LLMs required extensive training times ranging from 30 to 50 hours, whereas Transformer-based models trained from scratch completed training in under an hour. This stark contrast highlights the practicality of simpler Transformer-based models for clinical NLP tasks in resource-constrained environments. While adapters like GRN can enhance performance, their benefits are diminished by the substantial training times and limited improvements observed in scenarios with frozen weights and limited data.

As summarized in Table \ref{tab:camemBERT}, \ref{tab:alibert}, \ref{tab:drbert}, the experimental results compare different adapters for various biomedical pre-trained models (CamemBERT-bio, AliBERT, and DrBERT, respectively) based on accuracy, precision, recall, and F1 score, with full fine-tuning as the baseline. For CamemBERT-bio, the baseline achieved high performance across all metrics, while TinyAttention closely matched the baseline, and AdaptFormer and Lightweight showed slight reductions in recall. GRN achieved slightly higher recall than Lightweight. For AliBERT, the baseline exhibited strong performance, especially in precision. AdaptFormer and Lightweight had noticeable drops in recall but maintained high precision and accuracy. GRN provided balanced performance, and TinyAttention closely matched the baseline in accuracy and precision. For DrBERT, the baseline again delivered strong results. AdaptFormer and Lightweight showed decreased recall and F1 scores, while GRN demonstrated higher recall and comparable precision to the other adapters. TinyAttention matched the baseline in accuracy and precision with a slight decrease in recall. While full fine-tuning (baseline) provided the best performance, GRN adapters balanced performance and computational efficiency, making them suitable for scenarios with limited computational resources. However, there were no significant improvements when adapters were used to fine-tune the pre-trained model with limited data. In some cases, it degraded performance, as seen with the AdaptFormer and Lightweight adapters in AliBERT and DrBERT, respectively.

\begin{figure}[!htp]
	\centering
	\includegraphics[scale=0.45]{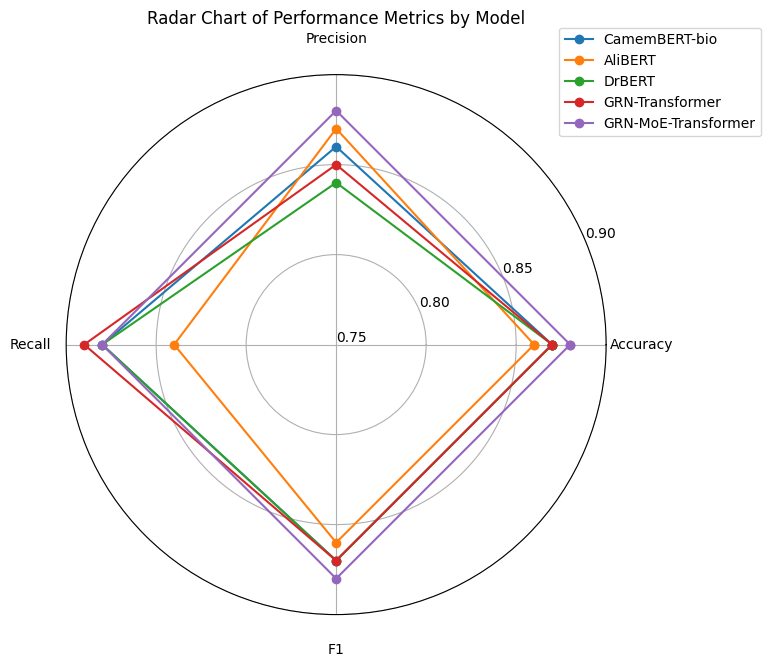}
	\caption{Performance comparison between biomedical pre-trained LLMs vs. Transformer-based models with different adapters.}
	\label{fig:all_compare}
\end{figure}

\begin{figure}[!htp]
	\centering
	\includegraphics[scale=0.55]{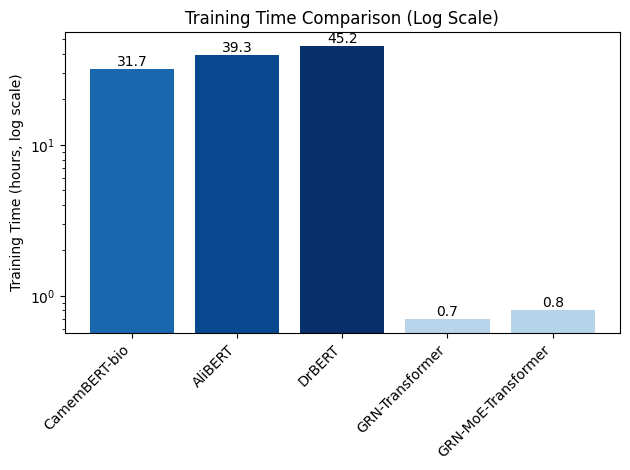}
	\caption{Training time comparison between biomedical pre-trained LLMs vs. Transformer-based models with different adapters.}
	\label{fig:training_time }
\end{figure}

The experimental results compare the performance of Transformers and MoE-Transformers using different adapters evaluated based on accuracy, precision, recall, and F1 score, as shown in Table \ref{tab:transformer}, and \ref{tab:moe_transformer}, respectively. The baseline models, trained from scratch without any adapters, provide a reference point against which the other models, also trained from scratch but with different adapters, are compared. These setups differ from using biomedical pre-trained models. From the results, two key points emerge. First, adapters help the Transformer, as all the adapters outperform the Transformer baseline. Second, with the more complex MoE-Transformer, adapters do not continually improve the MoE-Transformer baseline model; for instance, AdaptFormer and Lightweight show no significant improvement. However, both GRN and TinyAttention improve the MoE-Transformer compared to the baseline. Overall, GRN is the most effective technique, as it enhances the performance of both the Transformer and MoE-Transformer models.

Based on the results for biomedical pre-trained models and Transformer-based models trained from scratch, we compared the best performance of pre-trained models with GRN adapters to that of Transformer-based models with GRN adapters, as illustrated in Fig. \ref{fig:all_compare}. The results indicate no significant difference between fine-tuning the pre-trained models with adapters and applying adapters to Transformer-based models trained from scratch. This suggests that the advantage of using adapters for fine-tuning pre-trained models is unclear in scenarios with limited data. While adapters like GRN can improve model performance, their impact cannot distinguish between pre-trained models and those trained from scratch under data constraints. Overall, the benefit of employing adapters to fine-tune the pretrained LLM in limited data scenarios remains ambiguous.
    
\begin{figure}[!htp]
	\centering
	\includegraphics[scale=0.55]{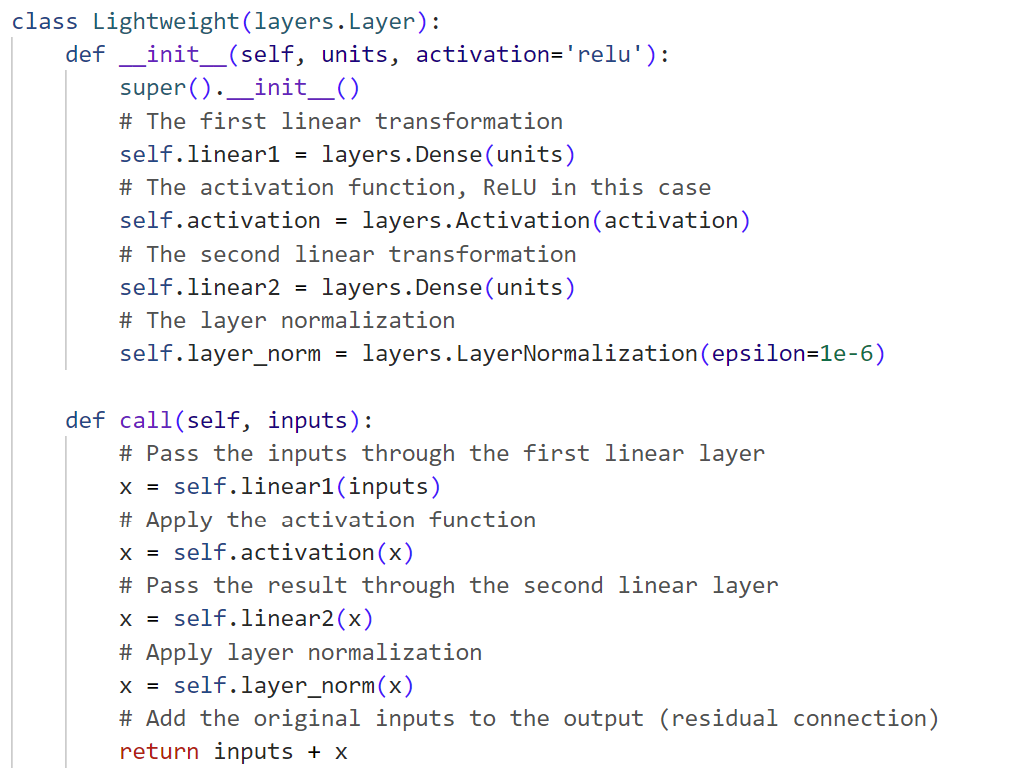}
	\caption{Pseudocode for Lightweight Adapter.}
	\label{fig:pseudo_lightweight}
 \vspace{-3mm}
\end{figure}

\begin{figure}[!htp]
	\centering
	\includegraphics[scale=0.55]{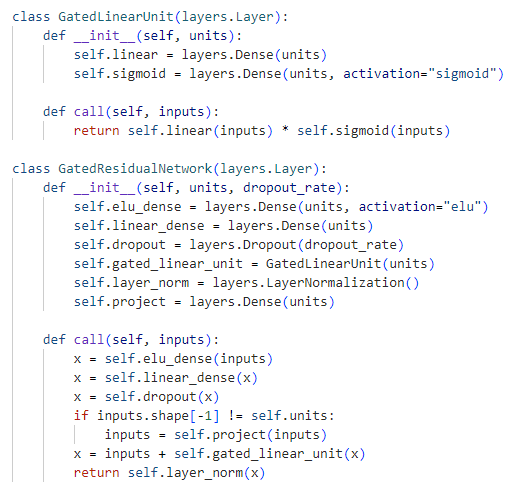}
	\caption{Pseudocode for GRN LoRA Adapter.}
	\label{fig:pseudo_grn}
 \vspace{-3mm}
\end{figure}

The GRN is designed to enhance neural networks' capabilities by integrating sophisticated gating mechanisms. At its core is the Gated Linear Unit, which combines a linear transformation with a sigmoid gated transformation, achieved through dense layers, and performs an element-wise multiplication of the linear and gated outputs. This mechanism ensures the network can dynamically control the information flow, enhancing its learning capabilities. Building upon this, the GatedResidualNetwork incorporates several key components: an ELU-activated dense layer that introduces non-linearity, a linear dense layer for further transformation, and a dropout layer to prevent overfitting. The gated linear unit is central to the GRN's function, which applies the gating mechanism to the residual connections. Additionally, layer normalization stabilizes and accelerates the training process, while a projection layer ensures that the input dimensionality matches the required units. Together, these elements form a robust architecture capable of effectively managing and transforming complex input data.

In contrast, the experimental results comparing training times for biomedical pre-trained models and Transformer-based models trained from scratch reveal significant differences. As shown in Fig. \ref{fig:training_time }, fine-tuning pre-trained models like CamemBERT-bio, AliBERT, and DrBERT with adapters takes substantially longer, ranging from 30 to 50 hours per experiment. In contrast, training transformer-based models with GRN adapters from scratch takes less than an hour. This highlights a crucial limitation of employing adapters for pre-trained models: the extensive computational capacity and training time required. Despite the adapters not showing significant performance improvements over training Transformer-based models from scratch, they demand significantly more computational resources and time. This makes using adapters in pre-trained models less appealing, especially in scenarios with limited computational resources and time constraints.

Our extensive experiments underscore the importance of carefully designing and implementing LoRA adapters when fine-tuning LLMs with limited data. These findings are consistent with recent research in other domains, such as programming and mathematics, as reported by \cite{biderman2024lora}. Their study shows that LoRA often underperforms full fine-tuning across various settings. Nonetheless, LoRA provides beneficial regularization, preserving the base model's performance on tasks outside the target domain more effectively than full fine-tuning. Moreover, LoRA offers stronger regularization compared to techniques like weight decay and dropout and supports maintaining more diverse outputs. Full fine-tuning tends to learn perturbations with a significantly higher rank (10-100 times) than typical LoRA configurations, which likely contributes to the performance differences observed. Consequently, it is crucial to exercise caution when applying LoRA adapters and fine-tuning pre-trained LLMs, particularly in sensitive domains like clinical NLP, where data privacy, limited data availability, and computational resource constraints are significant concerns. As McCoy et al. \cite{mccoy2024large} caution, inserting LLM-generated text directly into medical records could undermine communication, transparency, and the quality of healthcare, underscoring the need for caution in fine-tuning LLMs in clinical settings.

\section{Limitations and Future Works}
\label{sec:limitations}
While our study systematically evaluates adapter techniques under stringent compute (single Quadro P620) and data constraints (580,000 tokens), several limitations remain. First, due to our reliance on a single GPU, we lacked the memory capacity and compute throughput required to fine-tune large proprietary models like Deepseek and Grok \cite{liang5169443evaluation}. Replicating those experiments would have exceeded our hardware limits, both in terms of GPU memory and acceptable training time, making such evaluations infeasible within our study’s resource constraints. Second, we only compared fully trainable versus fully frozen backbones with adapters; intermediate freezing ratios (e.g., 40-80\% of layers frozen) may affect convergence speed and generalization, but were not explored. Third, our performance analysis focused primarily on accuracy and training time; other metrics such as memory footprint, latency, and energy consumption in diverse hospital server environments were not measured. Finally, we did not benchmark against the latest state‐of‐the‐art clinical NLP models (e.g., Mamba state‐space architectures \cite{gu2023mamba}) or other recent LoRA variants \cite{wang2025tina}, limiting our comparisons.

To address these gaps, future research should:
\begin{itemize} 
  \item Investigate hybrid fine‐tuning schemes that freeze varying ratios of the LLM layers while training the rest with adapters, evaluating how varying freeze ratios influence performance, convergence, and overfitting.
  \item Measure GPU memory, inference latency, and energy consumption on diverse hardware (e.g., A100) to quantify adapter trade-offs, and broaden our benchmarks to include cutting-edge LLMs for a full-spectrum clinical NLP evaluation.
  \item Benchmark our adapter and scratch‐trained Transformer models against emerging state‐space architectures such as Mamba and novel LoRA variants to position our findings within the current clinical NLP landscape.
\end{itemize}

\section{Conclusion}
\label{sec:conclusion}

Based on our comprehensive evaluation, this study concludes that employing adapter structures for fine-tuning biomedical pre-trained LLMs does not yield significant improvements in clinical NLP tasks under resource constraints. We found that simpler Transformer-based models trained from scratch perform comparably or better, especially in environments with limited computational resources and data availability. Among the adapter structures evaluated, the GRN demonstrated superior accuracy, precision, recall, and F1 score, making it the most effective adapter for enhancing clinical note classification. Furthermore, the stark contrast in training times - over 1000 hours for pre-trained LLMs versus under 6 hours for Transformer-based models - underscores the practicality of using simpler models in resource-constrained settings. This study contributes to the field by providing a viable solution for clinical NLP tasks in low-resource environments and identifying the GRN adapter as a practical approach to improve model performance without requiring extensive computational resources. Lastly, implementing the adapters with different algorithms is straightforward for reproducibility, as demonstrated by the pseudocode examples provided by Fig. \ref{fig:pseudo_lightweight} and \ref{fig:pseudo_grn}.

\section*{Acknowledgment}

The clinical data were provided by the Research Center at
CHU Sainte-Justine hospital. This work was supported in part by the Natural Sciences and Engineering Research Council (NSERC), in part by the Institut de Valorisation des données de l’Université de Montréal (IVADO), in part by the Fonds de la recherche en sante du Quebec (FRQS). Data and reproducible codes are available upon reasonable request to Prof. Philippe Jouvet, M.D., PhD. (Email: philippe.jouvet.med@ssss.gouv.qc.ca).

\bibliographystyle{IEEEtran}
\bibliography{IEEEabrv,Bibliography}

\begin{thebibliography}{10}
\providecommand{\url}[1]{#1}
\csname url@samestyle\endcsname
\providecommand{\newblock}{\relax}
\providecommand{\bibinfo}[2]{#2}
\providecommand{\BIBentrySTDinterwordspacing}{\spaceskip=0pt\relax}
\providecommand{\BIBentryALTinterwordstretchfactor}{4}
\providecommand{\BIBentryALTinterwordspacing}{\spaceskip=\fontdimen2\font plus
\BIBentryALTinterwordstretchfactor\fontdimen3\font minus
  \fontdimen4\font\relax}
\providecommand{\BIBforeignlanguage}[2]{{%
\expandafter\ifx\csname l@#1\endcsname\relax
\typeout{** WARNING: IEEEtran.bst: No hyphenation pattern has been}%
\typeout{** loaded for the language `#1'. Using the pattern for}%
\typeout{** the default language instead.}%
\else
\language=\csname l@#1\endcsname
\fi
#2}}
\providecommand{\BIBdecl}{\relax}
\BIBdecl

\bibitem{jozefowicz2016exploring}
R.~Jozefowicz, O.~Vinyals, M.~Schuster, N.~Shazeer, and Y.~Wu, ``Exploring the
  limits of language modeling,'' \emph{arXiv preprint arXiv:1602.02410}, 2016.

\bibitem{vaswani2017attention}
A.~Vaswani and et. al., ``Attention is all you need,'' \emph{Advances in Neural
  Information Processing Systems}, vol.~30, 2017.

\bibitem{radford2019language}
A.~Radford, J.~Wu, R.~Child, D.~Luan, D.~Amodei, I.~Sutskever \emph{et~al.},
  ``Language models are unsupervised multitask learners,'' \emph{OpenAI blog},
  vol.~1, no.~8, p.~9, 2019.

\bibitem{brown2020language}
T.~Brown, B.~Mann, N.~Ryder, M.~Subbiah, J.~D. Kaplan, P.~Dhariwal,
  A.~Neelakantan, P.~Shyam, G.~Sastry, A.~Askell \emph{et~al.}, ``Language
  models are few-shot learners,'' \emph{Advances in neural information
  processing systems}, vol.~33, pp. 1877--1901, 2020.

\bibitem{kaplan2020scaling}
J.~Kaplan, S.~McCandlish, T.~Henighan, T.~B. Brown, B.~Chess, R.~Child,
  S.~Gray, A.~Radford, J.~Wu, and D.~Amodei, ``Scaling laws for neural language
  models,'' \emph{arXiv preprint arXiv:2001.08361}, 2020.

\bibitem{achiam2023gpt}
J.~Achiam, S.~Adler, S.~Agarwal, L.~Ahmad, I.~Akkaya, F.~L. Aleman, D.~Almeida,
  J.~Altenschmidt, S.~Altman, S.~Anadkat \emph{et~al.}, ``Gpt-4 technical
  report,'' \emph{arXiv preprint arXiv:2303.08774}, 2023.

\bibitem{alammar2018illustrated}
J.~Alammar, ``The illustrated transformer,'' \emph{The Illustrated
  Transformer--Jay Alammar--Visualizing Machine Learning One Concept at a
  Time}, vol.~27, 2018.

\bibitem{bellani2016epidemiology}
G.~Bellani and et. al., ``Epidemiology, patterns of care, and mortality for
  patients with acute respiratory distress syndrome in intensive care units in
  50 countries,'' \emph{JAMA}, vol. 315, no.~8, pp. 788--800, 2016.

\bibitem{pediatric2015pediatric}
P.~A. L. I. C.~C. Group \emph{et~al.}, ``Pediatric acute respiratory distress
  syndrome: consensus recommendations from the pediatric acute lung injury
  consensus conference,'' \emph{Pediatric critical care medicine: a journal of
  the Society of Critical Care Medicine and the World Federation of Pediatric
  Intensive and Critical Care Societies}, p. 428, 2015.

\bibitem{sauthier2021estimated}
M.~Sauthier and et. al., ``Estimated pao2: A continuous and noninvasive method
  to estimate pao2 and oxygenation index,'' \emph{Critical care explorations},
  vol.~3, no.~10, 2021.

\bibitem{zaglam2014computer}
N.~Zaglam and et. al., ``Computer-aided diagnosis system for the acute
  respiratory distress syndrome from chest radiographs,'' \emph{Computers in
  biology and medicine}, vol.~52, pp. 41--48, 2014.

\bibitem{yahyatabar2020dense}
M.~Yahyatabar, P.~Jouvet, and F.~Cheriet, ``Dense-unet: a light model for lung
  fields segmentation in chest x-ray images,'' in \emph{2020 42nd Annual
  International Conference of the IEEE Engineering in Medicine \& Biology
  Society (EMBC)}.\hskip 1em plus 0.5em minus 0.4em\relax IEEE, 2020, pp.
  1242--1245.

\bibitem{le2021detecting}
T.~D. Le and et. al., ``Detecting of a patient's condition from clinical
  narratives using natural language representation,'' \emph{IEEE Open Journal
  of Engineering in Medicine and Biology}, vol.~3, pp. 142--149, 2022.

\bibitem{le2023adaptation}
T.-D. Le and et. al., ``Adaptation of autoencoder for sparsity reduction from
  clinical notes representation learning,'' \emph{IEEE Journal of Translational
  Engineering in Health and Medicine}, 2023.

\bibitem{lompo2024numerical}
B.~A. Lompo and T.-D. Le, ``Numerical attributes learning for cardiac failure
  diagnostic from clinical narratives-a lesa-camembert-bio approach,''
  \emph{arXiv preprint arXiv:2404.10171}, 2024.

\bibitem{lompo2024multi}
B.-A. Lompo and T.-D. Le, ``Multi-objective representation for numbers in
  clinical narratives using camembert-bio,'' \emph{arXiv preprint
  arXiv:2405.18448}, 2024.

\bibitem{le2023small}
T.-D. Le, P.~Jouvet, and R.~Noumeir, ``Improving transformer performance for
  french clinical notes classification using mixture of experts on a limited
  dataset,'' \emph{arXiv preprint arXiv:2303.12892}, 2024.

\bibitem{le2023grn}
T.-D. Le, ``{GRN-Transformer}: Enhancing motion artifact detection in {PICU}
  photoplethysmogram signals,'' \emph{arXiv preprint arXiv:2308.03722}, 2023.

\bibitem{fields2024survey}
J.~Fields, K.~Chovanec, and P.~Madiraju, ``A survey of text classification with
  transformers: How wide? how large? how long? how accurate? how expensive? how
  safe?'' \emph{IEEE Access}, 2024.

\bibitem{raiaan2024review}
M.~A.~K. Raiaan, M.~S.~H. Mukta, K.~Fatema, N.~M. Fahad, S.~Sakib, M.~M.~J.
  Mim, J.~Ahmad, M.~E. Ali, and S.~Azam, ``A review on large language models:
  Architectures, applications, taxonomies, open issues and challenges,''
  \emph{IEEE Access}, 2024.

\bibitem{singhal2023large}
K.~Singhal, S.~Azizi, T.~Tu, S.~S. Mahdavi, J.~Wei, H.~W. Chung, N.~Scales,
  A.~Tanwani, H.~Cole-Lewis, S.~Pfohl \emph{et~al.}, ``Large language models
  encode clinical knowledge,'' \emph{Nature}, vol. 620, no. 7972, pp. 172--180,
  2023.

\bibitem{zhangscaling}
B.~Zhang, Z.~Liu, C.~Cherry, and O.~Firat, ``When scaling meets llm finetuning:
  The effect of data, model and finetuning method,'' in \emph{The Twelfth
  International Conference on Learning Representations}.

\bibitem{chen2022revisiting}
G.~Chen, F.~Liu, Z.~Meng, and S.~Liang, ``Revisiting parameter-efficient
  tuning: Are we really there yet?'' in \emph{Proceedings of the 2022
  Conference on Empirical Methods in Natural Language Processing}, 2022.

\bibitem{hulora}
E.~J. Hu, P.~Wallis, Z.~Allen-Zhu, Y.~Li, S.~Wang, L.~Wang, W.~Chen
  \emph{et~al.}, ``Lora: Low-rank adaptation of large language models,'' in
  \emph{International Conference on Learning Representations}.

\bibitem{houlsby2019parameter}
N.~Houlsby, A.~Giurgiu, S.~Jastrzebski, B.~Morrone, Q.~De~Laroussilhe,
  A.~Gesmundo, M.~Attariyan, and S.~Gelly, ``Parameter-efficient transfer
  learning for nlp,'' in \emph{International conference on machine
  learning}.\hskip 1em plus 0.5em minus 0.4em\relax PMLR, 2019, pp. 2790--2799.

\bibitem{pfeiffer2020adapterhub}
J.~Pfeiffer, A.~R{\"u}ckl{\'e}, C.~Poth, A.~Kamath, I.~Vuli{\'c}, S.~Ruder,
  K.~Cho, and I.~Gurevych, ``Adapterhub: A framework for adapting
  transformers,'' in \emph{Proceedings of the 2020 Conference on Empirical
  Methods in Natural Language Processing: System Demonstrations}, 2020, pp.
  46--54.

\bibitem{chen2022adaptformer}
S.~Chen, C.~Ge, Z.~Tong, J.~Wang, Y.~Song, J.~Wang, and P.~Luo, ``Adaptformer:
  Adapting vision transformers for scalable visual recognition,''
  \emph{Advances in Neural Information Processing Systems}, vol.~35, pp.
  16\,664--16\,678, 2022.

\bibitem{le2021lightweight}
H.~Le, J.~Pino, C.~Wang, J.~Gu, D.~Schwab, and L.~Besacier, ``Lightweight
  adapter tuning for multilingual speech translation,'' in \emph{Proceedings of
  the 59th Annual Meeting of the Association for Computational Linguistics and
  the 11th International Joint Conference on Natural Language Processing
  (Volume 2: Short Papers)}, 2021, pp. 817--824.

\bibitem{zhao2022tiny}
H.~Zhao, H.~Tan, and H.~Mei, ``Tiny-attention adapter: Contexts are more
  important than the number of parameters,'' in \emph{Proceedings of the 2022
  Conference on Empirical Methods in Natural Language Processing}, 2022, pp.
  6626--6638.

\bibitem{le2024transformer}
T.-D. Le, ``Transformer meets gated residual networks to enhance
  photoplethysmogram artifact detection informed by mutual information neural
  estimation,'' \emph{arXiv preprint arXiv:2405.16177}, 2024.

\bibitem{kessler2017scattertext}
J.~Kessler, ``Scattertext: a browser-based tool for visualizing how corpora
  differ,'' in \emph{Proceedings of ACL 2017, System Demonstrations}, 2017, pp.
  85--90.

\bibitem{corr2306}
\BIBentryALTinterwordspacing
R.~Touchent, L.~Romary, and {\'{E}}.~de~la Clergerie, ``Camembert-bio: a tasty
  french language model better for your health,'' \emph{CoRR}, vol.
  abs/2306.15550, 2023. [Online]. Available:
  \url{https://doi.org/10.48550/arXiv.2306.15550}
\BIBentrySTDinterwordspacing

\bibitem{LabrakBDRMDG23}
\BIBentryALTinterwordspacing
Y.~Labrak, A.~Bazoge, R.~Dufour, M.~Rouvier, E.~Morin, B.~Daille, and
  P.~Gourraud, ``Drbert: {A} robust pre-trained model in french for biomedical
  and clinical domains,'' in \emph{Proceedings of the 61st Annual Meeting of
  the Association for Computational Linguistics (Volume 1: Long Papers), {ACL}
  2023, Toronto, Canada, July 9-14, 2023}, A.~Rogers, J.~L. Boyd{-}Graber, and
  N.~Okazaki, Eds.\hskip 1em plus 0.5em minus 0.4em\relax Association for
  Computational Linguistics, 2023, pp. 16\,207--16\,221. [Online]. Available:
  \url{https://doi.org/10.18653/v1/2023.acl-long.896}
\BIBentrySTDinterwordspacing

\bibitem{BerheDMMDZ23}
\BIBentryALTinterwordspacing
A.~Berhe, G.~Draznieks, V.~Martenot, V.~Masdeu, L.~Davy, and J.~Zucker,
  ``Alibert: {A} pre-trained language model for french biomedical text,'' in
  \emph{The 22nd Workshop on Biomedical Natural Language Processing and BioNLP
  Shared Tasks, BioNLP@ACL 2023, Toronto, Canada, 13 July 2023},
  D.~Demner{-}Fushman, S.~Ananiadou, and K.~Cohen, Eds.\hskip 1em plus 0.5em
  minus 0.4em\relax Association for Computational Linguistics, 2023, pp.
  223--236. [Online]. Available:
  \url{https://doi.org/10.18653/v1/2023.bionlp-1.19}
\BIBentrySTDinterwordspacing

\bibitem{macabiau2024label}
C.~Macabiau, T.-D. Le, K.~Albert, M.~Shahriari, P.~Jouvet, and R.~Noumeir,
  ``Label propagation techniques for artifact detection in imbalanced classes
  using photoplethysmogram signals,'' \emph{IEEE Access}, vol.~12, pp.
  81\,221--81\,235, 2024.

\bibitem{ding2023parameter}
N.~Ding, Y.~Qin, G.~Yang, F.~Wei, Z.~Yang, Y.~Su, S.~Hu, Y.~Chen, C.-M. Chan,
  W.~Chen \emph{et~al.}, ``Parameter-efficient fine-tuning of large-scale
  pre-trained language models,'' \emph{Nature Machine Intelligence}, 2023.

\bibitem{scikit-learn}
F.~Pedregosa and et. al, ``Scikit-learn: Machine learning in {P}ython,''
  \emph{Journal of Machine Learning Research}, vol.~12, pp. 2825--2830, 2011.

\bibitem{chollet2015keras}
F.~Chollet and et. al., ``keras,'' 2015.

\bibitem{popel2018training}
M.~Popel and et. al., ``Training tips for the transformer model,'' \emph{arXiv
  preprint arXiv:1804.00247}, 2018.

\bibitem{srivastava2014dropout}
N.~Srivastava and et. al., ``Dropout: a simple way to prevent neural networks
  from overfitting,'' \emph{The Journal of Machine Learning Research}, vol.~15,
  no.~1, pp. 1929--1958, 2014.

\bibitem{glorot2010understanding}
X.~Glorot and et. al., ``Understanding the difficulty of training deep
  feedforward neural networks,'' in \emph{Proceedings of the Thirteenth
  International Conference on Artificial Intelligence and Statistics}.\hskip
  1em plus 0.5em minus 0.4em\relax JMLR Workshop and Conference Proceedings,
  2010, pp. 249--256.

\bibitem{ioffe2015batch}
S.~Ioffe and et. al., ``Batch normalization: Accelerating deep network training
  by reducing internal covariate shift,'' in \emph{International Conference on
  Machine Learning}.\hskip 1em plus 0.5em minus 0.4em\relax PMLR, 2015, pp.
  448--456.

\bibitem{bjorck2018understanding}
N.~Bjorck and et. al., ``Understanding batch normalization,'' \emph{Advances in
  Neural Information Processing Systems}, vol.~31, 2018.

\bibitem{goutte2005probabilistic}
C.~Goutte and et. al., ``A probabilistic interpretation of precision, recall
  and f-score, with implication for evaluation,'' in \emph{European Conference
  on Information Retrieval}.\hskip 1em plus 0.5em minus 0.4em\relax Springer,
  2005, pp. 345--359.

\bibitem{biderman2024lora}
D.~Biderman, J.~G. Ortiz, J.~Portes, M.~Paul, P.~Greengard, C.~Jennings,
  D.~King, S.~Havens, V.~Chiley, J.~Frankle \emph{et~al.}, ``{LoRA} learns less
  and forgets less,'' \emph{arXiv preprint arXiv:2405.09673}, 2024.

\bibitem{mccoy2024large}
L.~G. McCoy, A.~K. Manrai, and A.~Rodman, ``Large language models and the
  degradation of the medical record,'' \emph{The New England journal of
  medicine}, 2024.

\bibitem{liang5169443evaluation}
J.~Liang, Y.~Lin, Z.~Guo, H.~Liang, J.~Ni, D.~Lin, J.~Qi, Z.~Huang, W.~Wang,
  and J.~He, ``Evaluation of llms accuracy and application in oncology
  principles and practice,'' \emph{Available at SSRN 5169443}.

\bibitem{gu2023mamba}
A.~Gu and T.~Dao, ``Mamba: Linear-time sequence modeling with selective state
  spaces,'' \emph{arXiv preprint arXiv:2312.00752}, 2023.

\bibitem{wang2025tina}
S.~Wang, J.~Asilis, {\"O}.~F. Akg{\"u}l, E.~B. Bilgin, O.~Liu, and
  W.~Neiswanger, ``Tina: Tiny reasoning models via {LoRA},'' \emph{arXiv
  preprint arXiv:2504.15777}, 2025.

\end{thebibliography}



\begin{IEEEbiography}[{\includegraphics[width=1in, height=1.25in, clip, keepaspectratio]{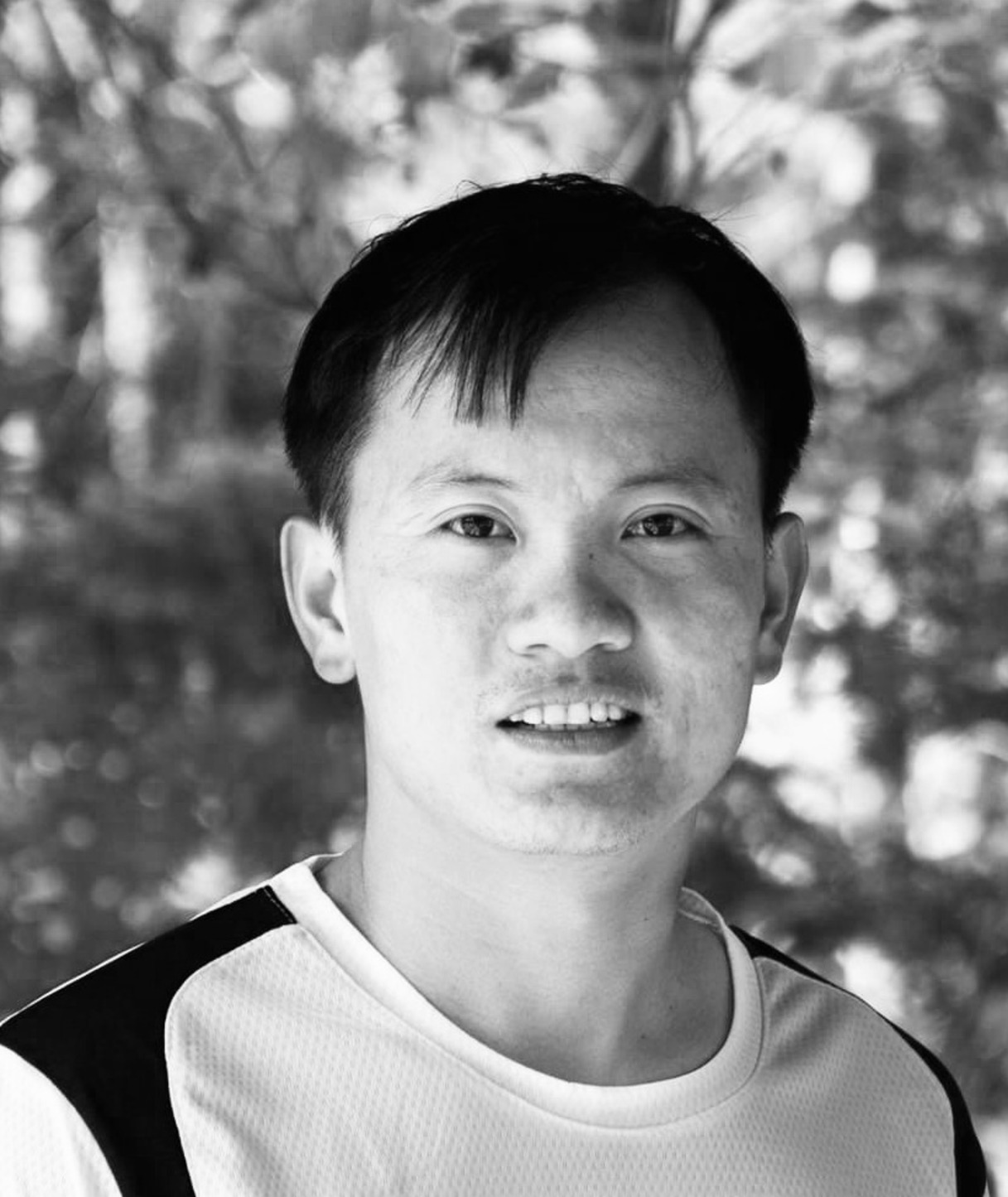}}]{Thanh-Dung Le} (Senior Member, IEEE) received a B.Eng. degree in mechatronics engineering from Can Tho University, Vietnam, an M.Eng. degree in electrical engineering from Jeju National University, S. Korea, and a Ph.D. in electrical engineering (Major in Applied Artificial Intelligence) from Ecole de Technologie Superieure (ETS), University of Quebec, Canada. From October 2023 to May 2024, he was a Postdoctoral Fellow with the Biomedical Information Processing Laboratory, ETS. His research interests include applied machine learning approaches for biomedical informatics problems. Before that, he joined the Institut National de la Recherche Scientifique, Canada, where he researched classification theory and machine learning. He is currently a Research Associate at the Interdisciplinary Center for Security, Reliability, and Trust (SnT) at the University of Luxembourg, focusing on applied machine learning approaches for satellite communications systems. He received the merit doctoral scholarship from Le Fonds de Recherche du Quebec Nature et Technologies. He also received the NSERC-PERSWADE fellowship in Canada and a graduate scholarship from the Korean National Research Foundation, S. Korea.
\vspace{-5mm}
\end{IEEEbiography}

\begin{IEEEbiography}[{\includegraphics[width=1in, height=1.25in, clip, keepaspectratio]{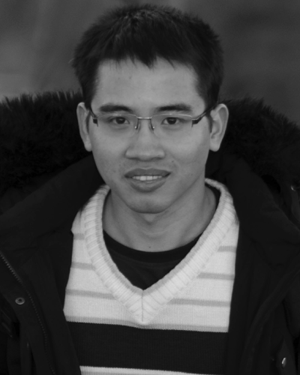}}]{Ti Ti Nguyen} (Member, IEEE) received the B.Eng. degree in electrical engineering from the Ho Chi Minh City University of Technology, Vietnam, in 2013, the M.Eng. degree in embedded system from the University of Rennes 1, France, in 2015, and the Ph.D. degree in telecommunications from Institut National de la Recherche Scientifique (INRS), Université du Québec, Canada, in 2020. Since 2020, he has been a Post-Doctoral Fellow with the Synchromedia, Ecole de Technologie Supérieure, Université du Québec. His current research interests include mobile edge computing and radio resource management, 5G new radio, MIMO, wireless security, and AI for wireless communications.
\vspace{-5mm}
\end{IEEEbiography}

\begin{IEEEbiography}[{\includegraphics[width=1in, height=1.25in, clip, keepaspectratio]{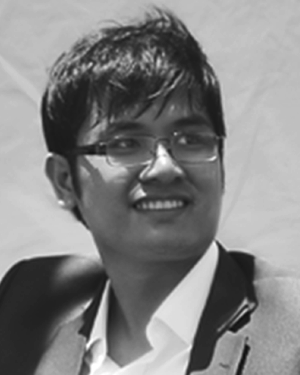}}]{Vu Nguyen Ha} (Senior Member, IEEE) received the B.Eng. degree (Hons.) from the French Training Program for Excellent Engineers in Vietnam, Ho Chi Minh City University of Technology, Vietnam, the Addendum degree from the École Nationale Supérieure des Télécommunications de Bretagne-Groupe des École des Télécommunications, Bretagne, France, in 2007, and the Ph.D. degree (Hons.) from the Institut National de la Recherche Scientifique-Énergie, Matériaux et Télécommunications, Université du Québec, Montreal, QC, Canada, in 2017. From 2016 to 2021, he worked as a Postdoctoral Fellow with the Ecole Polytechnique de Montreal, and then the Resilient Machine Learning Institute, École de Technologie Supérieure, University of Québec. He is currently a Research Scientist with the Interdisciplinary Centre for Security, Reliability, and Trust, University of Luxembourg. He was a recipient of the FRQNT Postdoctoral Fellowship for International Researcher (PBEEE) awarded by the Québec Ministry of Education, Canada, in 2018 and 2019. In 2021 and 2022, he was also awarded the Certificate for Exemplary Reviews by the IEEE Wireless Communications Letters.
\vspace{-5mm}
\end{IEEEbiography}

\begin{IEEEbiography}[{\includegraphics[width=1in, height=1.25in, clip, keepaspectratio]{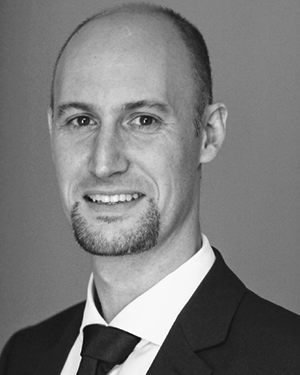}}]{Symeon Chatzinotas} (Fellow, IEEE) received the M.Eng. degree in telecommunications from the Aristotle University of Thessaloniki, Greece, in 2003, and the M.Sc. and Ph.D. degrees in electronic engineering from the University of Surrey, U.K., in 2006 and 2009, respectively. He is currently a Full Professor/Chief Scientist I and the Head of the Research Group SIGCOM, Interdisciplinary Centre for Security, Reliability and Trust, University of Luxembourg. In parallel, he is an Adjunct Professor with the Department of Electronic Systems, Norwegian University of Science and Technology and a Collaborating Scholar with the Institute of Informatics and Telecommunications, National Center for Scientific Research “Demokritos.” In the past, he has lectured as a Visiting Professor with the University of Parma, Italy and contributed in numerous research and development projects for the Institute of Telematics and Informatics, Center of Research and Technology Hellas and the Mobile Communications Research Group, Center of Communication Systems Research, University of Surrey. He has authored more than 700 technical papers in refereed international journals, conferences, and scientific books and has received numerous awards and recognitions, including the IEEE Fellowship and an IEEE Distinguished Contributions Award. He is currently on the editorial board of the IEEE Transactions on Communications, IEEE Open Journal of Vehicular Technology, and the International Journal of Satellite Communications and Networking.
\vspace{-5mm}
\end{IEEEbiography}

\begin{IEEEbiography}
[{\includegraphics[width=1in,height=1.25in, clip, keepaspectratio]{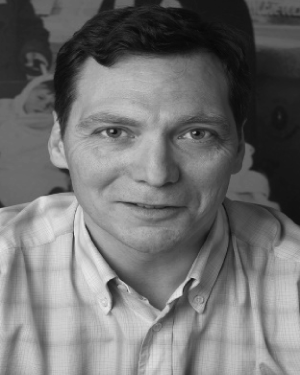}}]{Philippe Jouvet}received the M.D. degree from Paris V University, Paris, France, in 1989, the M.D. specialty in pediatrics and the M.D. subspecialty in intensive care from Paris V University, in 1989 and 1990, respectively, and the Ph.D. degree in pathophysiology of human nutrition and metabolism from Paris VII University, Paris, in 2001. He joined the Pediatric Intensive Care Unit of Sainte Justine Hospital—University of Montreal, Montreal, QC, Canada, in 2004. He is currently the Deputy Director of the Research Center and the Scientific Director of the Health Technology Assessment Unit, Sainte Justine Hospital–University of Montreal. He has a salary award for research from the Quebec Public Research Agency (FRQS). He currently conducts a research program on computerized decision support systems for health providers. His research program is supported by several grants from the Sainte-Justine Hospital, Quebec Ministry of Health, the FRQS, the Canadian Institutes of Health Research (CIHR), and the Natural Sciences and Engineering Research Council (NSERC). He has published more than 160 articles in peer-reviewed journals. Dr. Jouvet gave more than 120 lectures in national and international congresses.
\vspace{-5mm}
\end{IEEEbiography}

\begin{IEEEbiography}[{\includegraphics[width=1in,height=1.25in, clip, keepaspectratio]{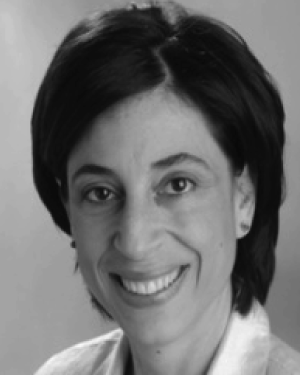}}]{Rita Noumeir} (Member, IEEE) received master's and Ph.D. degrees in biomedical engineering from École Polytechnique of Montreal. She is a Full Professor at the Department of Electrical Engineering, École de Technologie Superieure, University of Quebec, Canada. She is a Research Chair in artificial intelligence (AI) in healthcare/ Digital health and life sciences from the Fonds de Recherche du Québec- Santé (FRQS). Concurrently, she is the Co-Director of the research cluster on AI Applied to Acute Child Care, FRQS. Her main research interest is applying AI methods to create decision support systems. She has extensively worked in healthcare information technology and image processing. She has also provided consulting services in large-scale software architecture, healthcare interoperability, workflow analysis, and technology assessment for several international software and medical companies, including Canada Health Infoway.
\vspace{-5mm}
\end{IEEEbiography}

\EOD

\end{document}